%% file: example.tex
\documentclass{article}

\usepackage[preprint]{corl_2023} % Uncomment for pre-prints (e.g., arxiv); This is like ``final'', but will remove the CORL footnote.

\usepackage{times}
\usepackage{epsfig}
\usepackage{graphicx}
\usepackage{amsmath}
\usepackage{amssymb}
\usepackage{booktabs}
\usepackage{balance}
\usepackage{multirow}
\usepackage{multicol}
\usepackage{colortbl}
\definecolor{mygray}{gray}{.85}
\definecolor{myhighlight}{RGB}{193,210,240}
\usepackage{mwe}
\usepackage{authblk}
\usepackage{algorithmic,float}
\usepackage[linesnumbered,ruled,vlined,noline]{algorithm2e}
\usepackage{etoolbox}
\usepackage{relsize}
\usepackage{comment}
\usepackage{bm}
\usepackage{enumitem}
\usepackage{adjustbox}
\usepackage{soul}
\usepackage{wrapfig}
\usepackage{array}
\usepackage{hyperref}
\renewcommand\arraystretch{1.2}

\input{mlVecMat}

\usepackage{xcolor}

\title{Embodied Task Planning with Large Language Models}

\author{ \textbf{Zhenyu Wu}\textsuperscript{1}\textbf{,} \textbf{~Ziwei Wang}\textsuperscript{2,3}\textbf{,} \textbf{~Xiuwei Xu}\textsuperscript{2,3}\textbf{,} \textbf{~Jiwen Lu}\textsuperscript{2,3}\textbf{,} \textbf{~Haibin Yan}\textsuperscript{1}\thanks{Corresponding author.}\\
\vspace{-0.3cm}
\textsuperscript{1}School of Automation, Beijing University of Posts and Telecommunications, China\\
~\textsuperscript{2}Department of Automation, Tsinghua University, China \\
~\textsuperscript{3}Beijing National Research Center for Information Science and Technology, China \\
{\tt\small \{wuzhenyu, eyanhaibin\}@bupt.edu.cn; yali110136@gmail.com;}\\
{\tt\small xxw21@mails.tsinghua.edu.cn; lujiwen@tsinghua.edu.cn}\\
{\href{https://gary3410.github.io/TaPA}{https://gary3410.github.io/TaPA}} \\
}

\begin{document}
\maketitle
%%%%%%%%% ABSTRACT
\vspace{-0.4cm}
\begin{abstract}
  \input{./tex/abstract}
\end{abstract}


%%%%%%%%% BODY TEXT
\section{Introduction}
  \input{./tex/introduction}
  
\section{Related Work}
  \input{./tex/related}

\section{Approach}
  \input{./tex/approach}

\section{Experiment}
  \input{./tex/experiment}

\section{Conclusion}
  \input{./tex/conclusion}
\bibliography{example}  % .bib

\newpage
\appendix
\section*{Supplementary Material}
\input{./tex/supp}

\end{document}

%% file: mlVecMat.tex
\usepackage{amsmath}
\usepackage{amsfonts}
\usepackage{amssymb}
\usepackage{wrapfig}
\usepackage{subcaption}
\usepackage{multirow}
 \usepackage{mathtools} 

\usepackage{verbatim}

\usepackage{anyfontsize}

\usepackage{microtype}
\usepackage{graphicx}
\usepackage{booktabs} % for professional tables
\usepackage{multirow}
\usepackage{amsmath,amssymb}
\usepackage{booktabs}
\usepackage{caption,subcaption}

\usepackage{xcolor}
\definecolor{mygreen}{HTML}{3cb44b}
\definecolor{skyblue}{HTML}{beffff}
\definecolor{lightgreen}{HTML}{90ee90}

\usepackage{color, colortbl}

\definecolor{emerald}{rgb}{0.31, 0.78, 0.37}

\usepackage{tcolorbox}
\usepackage{enumitem}
\setitemize{itemsep=10pt,topsep=0pt,parsep=0pt,partopsep=0pt}
\usepackage{colortbl}

\usepackage{xcolor}
\definecolor{mygreen}{HTML}{3cb44b}
\colorlet{myyellow}{green!10!orange!90!}
\makeatletter

\usepackage{tikz}
\usetikzlibrary{arrows,shapes,snakes,automata,backgrounds,fit,petri}
\usepackage{adjustbox}

\newcommand{\RN}[1]{%
	\textup{\lowercase\expandafter{\it \romannumeral#1}}%
}
\usepackage{tabu}
\newcommand{\beq}{\vspace{0mm}\begin{equation}}
\newcommand{\eeq}{\vspace{0mm}\end{equation}}
\newcommand{\beqs}{\vspace{0mm}\begin{eqnarray}}
\newcommand{\eeqs}{\vspace{0mm}\end{eqnarray}}
\newcommand{\barr}{\begin{array}}
\newcommand{\earr}{\end{array}}

\usepackage{color, colortbl}
\definecolor{Gray}{gray}{0.93}

\usepackage{lipsum}

\usepackage{pifont}% http://ctan.org/pkg/pifont

\usepackage{makecell}

\usepackage{xcolor,amsmath}
\usepackage[linesnumbered,ruled,vlined]{algorithm2e}
\DontPrintSemicolon

\usepackage{xcolor}
\definecolor{mygreen}{HTML}{3cb44b}

\SetKwComment{Comment}{\color{green!50!black}\# }{}

\SetKwProg{Function}{def}{:}{}

\SetKwProg{For}{for}{:}{}
\SetKwProg{If}{if}{:}{}
\newcommand{\VarSty}[1]{\textnormal{\ttfamily\color{blue!90!black}#1}\unskip}

%% file: tex/abstract.tex
Equipping embodied agents with commonsense is important for robots to successfully complete complex human instructions in general environments. Recent large language models (LLM) can embed rich semantic knowledge for agents in plan generation of complex tasks, while they lack the information about the realistic world and usually yield infeasible action sequences. In this paper, we propose a TAsk Planing Agent (TaPA) in embodied tasks for grounded planning with physical scene constraint, where the agent generates executable plans according to the existed objects in the scene by aligning LLMs with the visual perception models. Specifically, we first construct a multimodal dataset containing triplets of indoor scenes, instructions and action plans, where we provide the designed prompts and the list of existing objects in the scene for GPT-3.5 to generate a large number of instructions and corresponding planned actions. The generated data is leveraged for grounded plan tuning of pre-trained LLMs. During inference, we discover the objects in the scene by extending open-vocabulary object detectors to multi-view RGB images collected in different achievable locations. Experimental results show that the generated plan from our TaPA framework can achieve higher success rate than LLaVA and GPT-3.5 by a sizable margin, which indicates the practicality of embodied task planning in general and complex environments.
% Our code and demo are available at \href{https://gary3410.github.io/TaPA}{https://gary3410.github.io/TaPA}.

%% file: tex/introduction.tex
Equipping embodied agents with general commonsense knowledge to accomplish complex tasks based on the natural language commands is desirable in many applications such as domestic service \cite{wu2023tidybot}, medical treatment \cite{li2023llava, zhao2023chatcad+, sun2023pathasst} and agricultural picking \cite{vemprala2023chatgpt, stella2023can}. 
Due to the limited training samples and diverse tasks in downstream applications, directly training an embodied agent across different deployment scenarios is infeasible. Recent progress in large language models (LLMs) \cite{touvron2023llama,zhang2023llama,peng2023instruction, zhu2023minigpt} acquires rich commonsense knowledge from the vast web data, whose knowledge can be potentially leveraged by embodied agents to generate action plans for human requirements represented in natural language. 

However, LLMs cannot perceive the surrounding scenes and may generate inexecutable actions due to the requirement of interacting with non-existed objects. 
For example, given the human command "Give me some wine", the generated action steps from GPT-3.5 are "pouring wine from the bottle to the glass". There may be only mugs instead of glasses in the realistic scenes, and the executable actions should be "pouring wine from the bottle to the mug". Therefore, grounding the task plan generated by LLMs to the physical world is necessary to construct embodied agents for complex task accomplishment.

To acquire executable task plans in the given physical scenes, many previous works filter or align the generated actions by considering the visual clues in the scene for the task of general manipulation of tabletop objects \cite{shridhar2022cliport,nair2022r3m,jang2022bc}. In order to further diversify tasks in house-level environments, SayCan \cite{brohan2023can} and LLM-Planner \cite{song2022llm} employ visual navigation to collect information in the house for the challenging grounded plan generation. Nevertheless, SayCan can only accomplish tasks in the kitchen scenarios and LLM-Planner performs planning in the ALFRED simulator \cite{shridhar2020alfred} where most tasks are simple such as putting and placing. They both fail to satisfy the requirement of numerous complex tasks and diverse deployment scenarios in our daily life.

% \begin{figure}[t]
%   \centering
%   \includegraphics[width=1\textwidth]{figs/figure_1_v6.png}
%   \caption{The pipeline of our embodied task planning framwork. We first collect multiple RGB images in different achivable standing points and views, and utilize an open-vocabulary detector to generate the list of existing objects in the scene. With the human instructions and the predicted object list, our TaPA can generate executable action plans for subsequent navigation or manipulation.}
  
%   \label{fig:pipeline}
%   \vspace{-0.8cm}
% \end{figure}

\begin{figure}[t]
  \centering
  \includegraphics[width=1\textwidth]{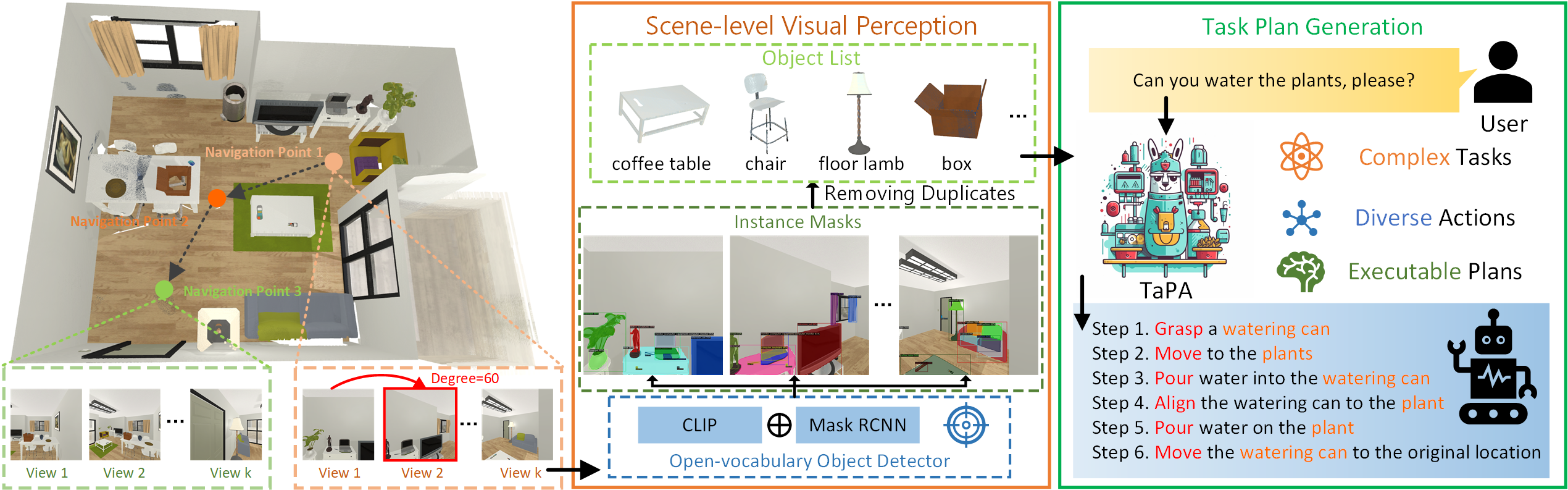}
  \caption{Our embodied task planning framework collects multiple RGB images from various standing points and viewpoints. Utilizing an open vocabulary detector generates a list of objects existed in the scene. Combining human instructions and the predicted object list, our TaPA generates executable action plans for navigation or manipulation tasks.}
  
  \label{fig:pipeline}
  \vspace{-2cm}
\end{figure}

In this paper, we present a task planning agent called TaPA for embodied task plan grounding in physical scenes. The unreleased SayCan cannot be applied in diverse indoor scenarios, and the LLM-Planner in the ALFRED benchmark fails to generate plans for complex tasks due to the pre-defined simple instructions in the simulator. 
On the contrary, our agent can generate grounded plans without constraining task types and target objects. 
Therefore, Our agent acquires general commonsense knowledge to yield action steps for complex household tasks such as making sandwiches and setting tables, which provides the foundational instructions for the downstream navigation and manipulation process to deal with high-level requirements from humans. Figure \ref{fig:pipeline} demonstrates the overall pipeline of our TaPA that generates the executable action steps by considering the scene information and the human instructions. 
Figure \ref{fig:data_comparison} shows the statistical difference between our TaPA and conventional ALFRED benchmark, where our tasks are much more complex with longer steps for accomplishment. More specifically, we first construct a multimodal dataset where each sample is a triplet of visual scenes, instructions, and corresponding plans. 
%We represent the visual scenes by the list of existing objects and carefully design the text prompt for GPT-3.5 to generate 15K instructions and corresponding action plans that are grounded to the physical scenes. 
By leveraging the generated dataset, we finetune the pre-trained LLaMA \cite{touvron2023llama} network by predicting the action steps based on the object list of the scene, which is employed as our task planner. For the acquisition of the object list during inference, the embodied agent effectively visits standing points to collect RGB images providing sufficient information in different views, and generalizes the open-vocabulary detector for multi-view images to acquire the list of existed objects. Our TaPA agent achieves higher success rate of the generated action plans compared with the state-of-the-art LLMs including LLaMA and GPT-3.5 and large multimodal models (LMMs) such as LLaVA \cite{liu2023visual}. Our contributions can be summarized as follows:

\begin{itemize}[leftmargin=*]
    \item To the best of our knowledge, we propose the first benchmark for complex embodied task planning that is practical in realistic indoor deployment scenarios.
    \item We design a framework for large-scale multimodal dataset generation in order to train the task planner from pre-trained LLMs and construct a multimodal dataset containing 80 indoor scenes with 15K instructions and corresponding action plans.
    \item We evaluate different LLMs and LMMs for complex embodied task planning in our benchmark, and conduct the ablation study to select the optimal representation of visual scenes for executable action generation.

\end{itemize}

%% file: tex/related.tex
\textbf{Large pre-trained models:} Large-scale pre-trained models have revolutionized the natural language processing (NLP) \cite{brown2020language,kenton2019bert,hu2021lora} and the computer vision \cite{kirillov2023segment,bangalath2022bridging,zhou2022detecting} communities in recent years. Benefiting from the vast training data and numerous parameters, the large pre-trained models acquire strong generalization ability across different deployment scenarios. For large language models, recent studies show that they not only perform well in NLP tasks, but also emerge the ability to master the rich knowledge about the realistic world with factual answers. Therefore, LLMs such as LLaMA \cite{touvron2023llama}, GPT-3 \cite{floridi2020gpt} are widely adopted to diverse tasks by interacting with input from other modalities such as visual feature learning \cite{li2023blip,sammani2022nlx}, pseudo-code generation \cite{bubeck2023sparks}, tool usage \cite{schick2023toolformer} and math problem solving \cite{zong2022solving}. For large vision models, objects in the open environments can be detected \cite{zhou2022detecting, zang2023contextual} or segmented \cite{ghiasi2021open} for scene understanding, where bounding boxes and masks are generated for all scenarios and visual features are aligned with text embedding for category assignment. To learn the joint embedding space of language and vision for multimodal tasks, CLIP \cite{radford2021learning} leverages contrastive learning to minimize the distance between similar image-text pairs. LLaVA \cite{liu2023visual} synthesized a multimodal dataset with images, captions and bounding boxes in the tasks of conversation, detailed description and complex reasoning, so that the instructional tuning of LLMs acquires general-purpose instruction-following visual agent.
In this paper, we leverage LLMs to generate executable plans for embodied tasks with the visual information acquired from the open-vocabulary detection models.

% \begin{figure}[ht]
% \centering
% 	\subfloat[subcaption1]{\includegraphics[width = 0.3\textwidth]{figs/alfred.png}}
% 	\quad
% 	\subfloat[subcaption2]{\includegraphics[width = 0.3\textwidth]{figs/tapa_v4.png}}
% 	\quad
% 	\subfloat[subcaption3]{\includegraphics[width = 0.3\textwidth]{figs/action_step.png}} 
% \caption{Comparison of TaPA and ALFRED dataset composition. 
%  The pie chart shows the top 20 frequently appearing verbs (inner circle) and the corresponding top 4 nouns (outer circle), where the number beside the verbs mean the average steps to complete the task. 
%  The bar chart shows the percentage of instructions with different numbers of actions, TaPA contains more complex instructions compared to ALFRED.}
% \label{fig:label}
% \end{figure}

\begin{figure}[t]
\setlength{\abovecaptionskip}{-0.1cm}
\setlength{\belowcaptionskip}{-0.1cm}
 \centering
 \includegraphics[width=1\textwidth]{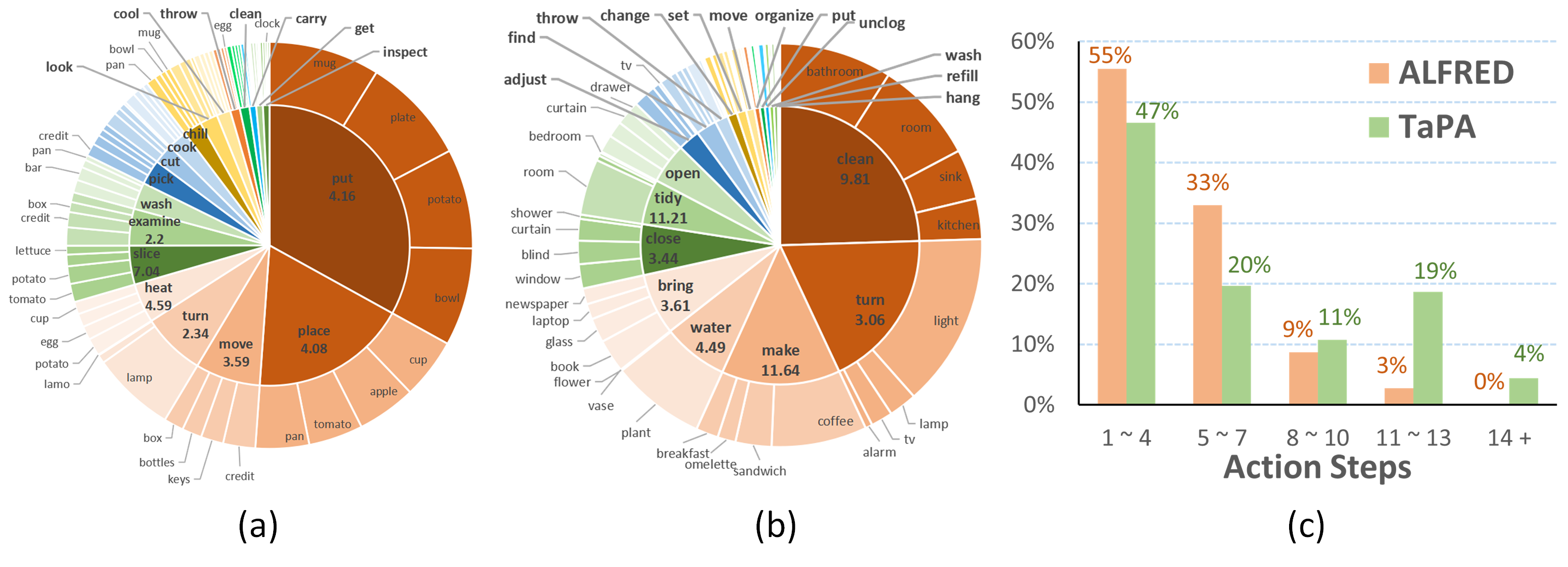}
 \caption{Statistical comparison of TaPA and ALFRED dataset. 
 The pie chart shows the top 20 frequently appearing verbs (inner circle) and the corresponding top 4 nouns (outer circle) for each verb. The bar chart shows the percentage of instructions with different numbers of implementation actions, where TaPA contains more complex instructions compared to ALFRED.}
 \label{fig:data_comparison}

\end{figure}

\textbf{Language model grounding for embodied tasks: }
An embodied agent not only requires active exploration \cite{wu2022smart}, manipulation \cite{liu2022ge}, and scene perception \cite{xu2023dspdet3d, xu2022back} as well as embodied task planning ability.
Embodied task planning aims to generate executable action steps in the given environments, where action plans are generated from grounded LLMs by receiving information from the surrounding environments \cite{blukis2022persistent,zellers2021piglet,akakzia2020grounding} or prompt engineering \cite{huang2022language}. For the former, agents acquire the feedback from environments by interacting with the objects to ground the task plan. Li \emph{et al.} \cite{li2022pre} employed LLMs as a general scaffold for interactive decision-making in complex tasks, where the generated policies were grounded to the given environments for executable implementation according to the action feedback. For prompt engineering, researchers carefully design the language prompts for LLMs to guide them to ground the generated content. Huang \emph{et al.} \cite{huang2022language} prompted simple examples of task instructions and corresponding actions for LLMs to produce plausible task plans, and filtered out executable subsets by constructing mapping with semantic similarity. To enable the LLMs to be aware of the surrounding scenes with boosted plan plausibility, Brohan \emph{et al.} \cite{brohan2023can} and Song \emph{et al.} \cite{song2022llm} extracted the visual information of the scene by latent features or object names for LLMs, where the generated plans were limited to the one with the highest success rate for task completion. However, these works can only accomplish simple tasks such as placing and putting in the VirtualHome \cite{puig2018virtualhome} or ALFRED simulators, which fail to be applied to practical deployment scenarios with diverse complex tasks.

%% file: tex/approach.tex
In this section, we first describe the construction of the multimodal instruction dataset that is leveraged to tune our TaPA task planner, and then describe the details of grounding embodied task plans to the visual scene with image collection and open-vocabulary detection.

\subsection{Data Generation of Embodied Task Planning}
Although large vision-language models (VLM)~\cite{liu2023visual, li2023otter} and large multimodal models \cite{lyu2023macaw, ye2023mplug, gao2023assistgpt, zhao2023chatbridge, chen2023x} have achieved surprising performance on a wide range of complex perception tasks, embodied task planning that is grounded to the realistic indoor scenes still remains challenging due to the lack of the large-scale multimodal dataset to train the planning agent. Considering the recent success of GPT models on high-level human-like reasoning, we leverage GPT-3.5 with the presented scene representation and designed prompt to generate the large-scale multimodal dataset for our planning agent tuning.

\begin{wrapfigure}{r}{6.5cm}
  \centering
  \includegraphics[width=0.47\textwidth]{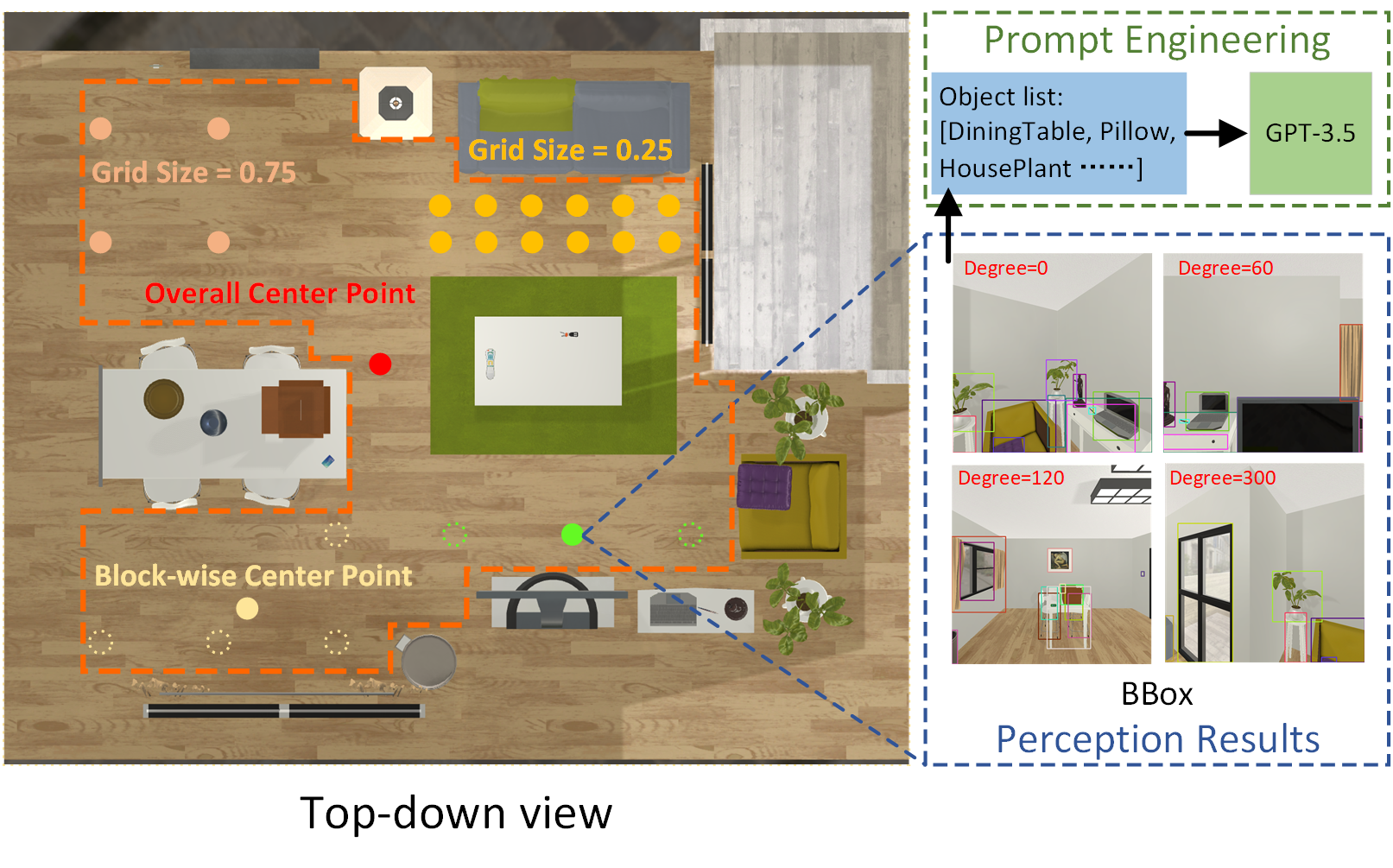}
  \caption{The pipeline of embedding the scene information for the LLM to generate executable actions. The image collection strategies to acquire the list of all existed objects in the scene include random sampling, traversal sampling, the overall center point and block-wise center points, where the object list is leveraged as the condition for action planning. The dashed circles in different colors represent grids in various blocks for block-wise center point selection.}
  \label{fig:data_generation}
\end{wrapfigure}

% Given an embodied 3D scene $X_v$, we directly extract all categories of objects from the ground truth and unify them into a list $X_l$, e.g., $[table, chair, keyboard,...]$. 
% 
Given an embodied 3D scene $X_s$, we directly utilize the class names of all objects as the representation of the scene which is denoted as $X_l$. All duplicate names are removed to provide scene information for the LLM such as $X_l=[table, chair, keyboard,...]$. Based on the above scene information, a simple approach used in ALFRED benchmark \cite{shridhar2020alfred} to generate the multimodal instruction following the dataset for embodied task plans is to artificially design a series of instructions with corresponding step-by-step actions. However, the hand-crafted design requires extremely high annotation cost to generate complex task plans that are practical for realistic service robots such as tidying up the bathroom and making sandwiches. To efficiently generate the large-scale complex instructions $X_q$ and executable corresponding plans $X_a$ for the given 3D scene, we design a prompt to simulate the scenarios of embodied task planning for GPT-3.5 to automatically synthesize data based on the object name list $X_l$. As shown in Table \ref{tab:prompt} of the supplementary materials, our prompt describes the definition of embodied task planning, the requirements and several examples of generated instructions and corresponding action plans. Specifically, the prompt designs a conversation between the service robot and humans to generate executable instructions and actions, which simulates the exploration of robots in the embodied environments and provides the requirements from humans. The generated instructions are diverse including requests, commands and queries, where only instructions with explicitly executable actions are added to our dataset. Meanwhile, we emphasize that the target object of the generated action should be constrained within the object list $X_l$ to mitigate the object hallucination that leads to inexecutable plans. For the object list leveraged in the prompt for dataset generation, we directly utilize the groundtruth label of existed instances in the scene. In Table \ref{tab:example}, we show examples of the generated sample containing the object name list of the scene, the instruction and the executable action steps. In embodied task planning, the agent can only get access to the visual scene containing all interactive objects without the groundtruth object list. Therefore, we construct the multimodal dataset by defining triplets for each sample as $\bm{X}=(X_v, X_q, X_a)$. For the training stage of the task planner, we directly leverage the groundtruth object list for each scene to avoid the influence of inaccurate visual perception. For the inference phase, the extended open-vocabulary object detector predicts the list of all existed objects in the scene.

We employ the AI2-THOR simulator \cite{kolve2017ai2} as the embodied environment for our agent, where we split the scenes with 80 for training and 20 for evaluation. To enlarge the scale and diversity of instructions and action steps in training samples for effective task planner finetuning, we expand the original 80 training scenes to 6400 training scenes by directly modifying the groundtruth object list. For each scene type, we initially acquire the list of objects that possibly appear in this type of scene by enumerating all rooms in the same room type. Then we randomly substitute existed objects with other ones that possibly exist in the same room type and are not observed. The plausibility constraint aims to prevent generating counterintuitive objects for given scene types. We collected 15K samples for training and leverages another 60 triplets for evaluation with our multimodal data generation framework.

\begin{table*}[t!]\centering
\begin{minipage}{1.0\columnwidth}\vspace{0mm}    \centering
\begin{tcolorbox} 
    \centering
   
    %  \hspace{-10mm}
      \footnotesize
    \begin{tabular}{p{0.46\columnwidth} p{0.2\columnwidth} p{0.3\columnwidth}}

\includegraphics[height=6cm]{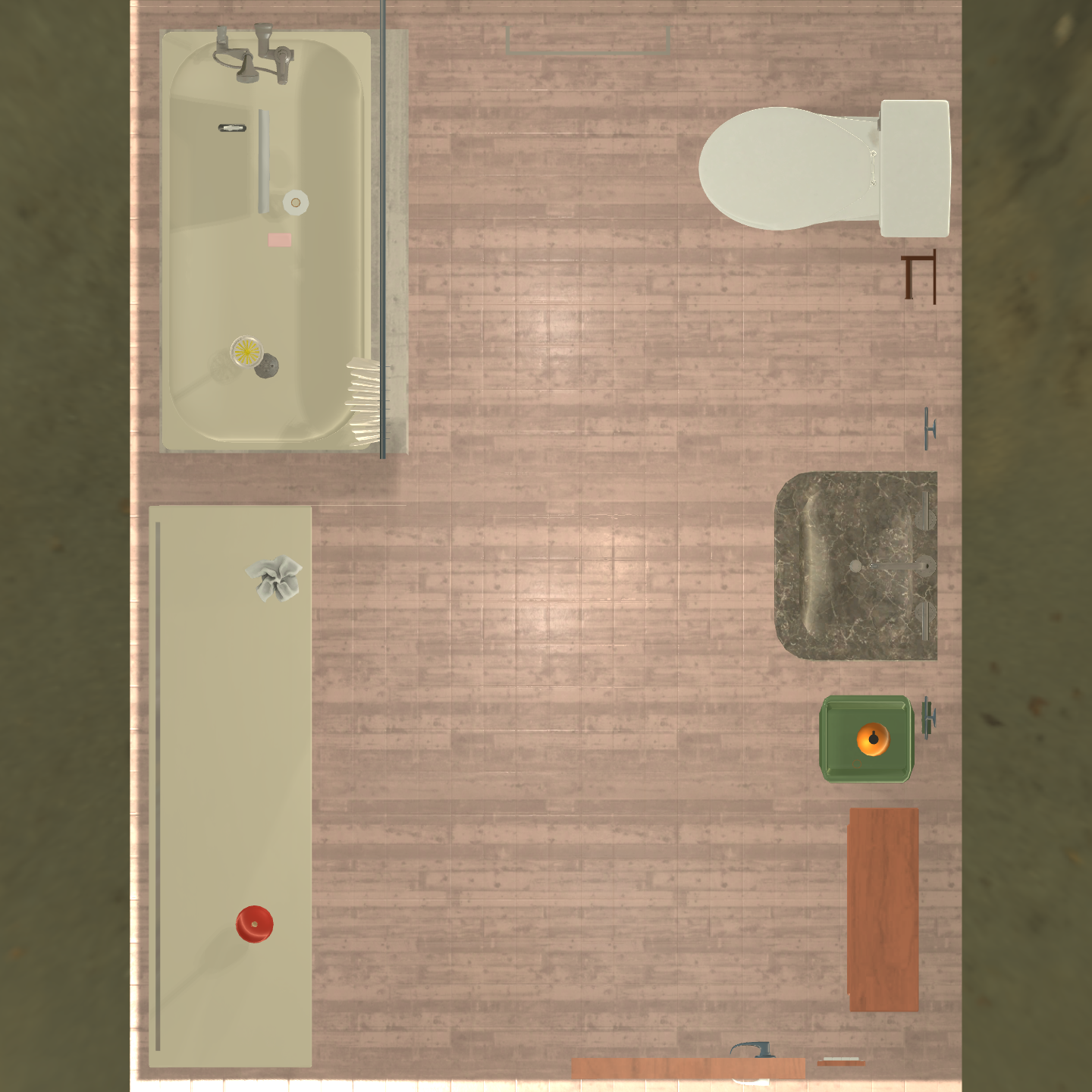} & \multicolumn{2}{p{0.5\columnwidth}}{\includegraphics[height=6cm]{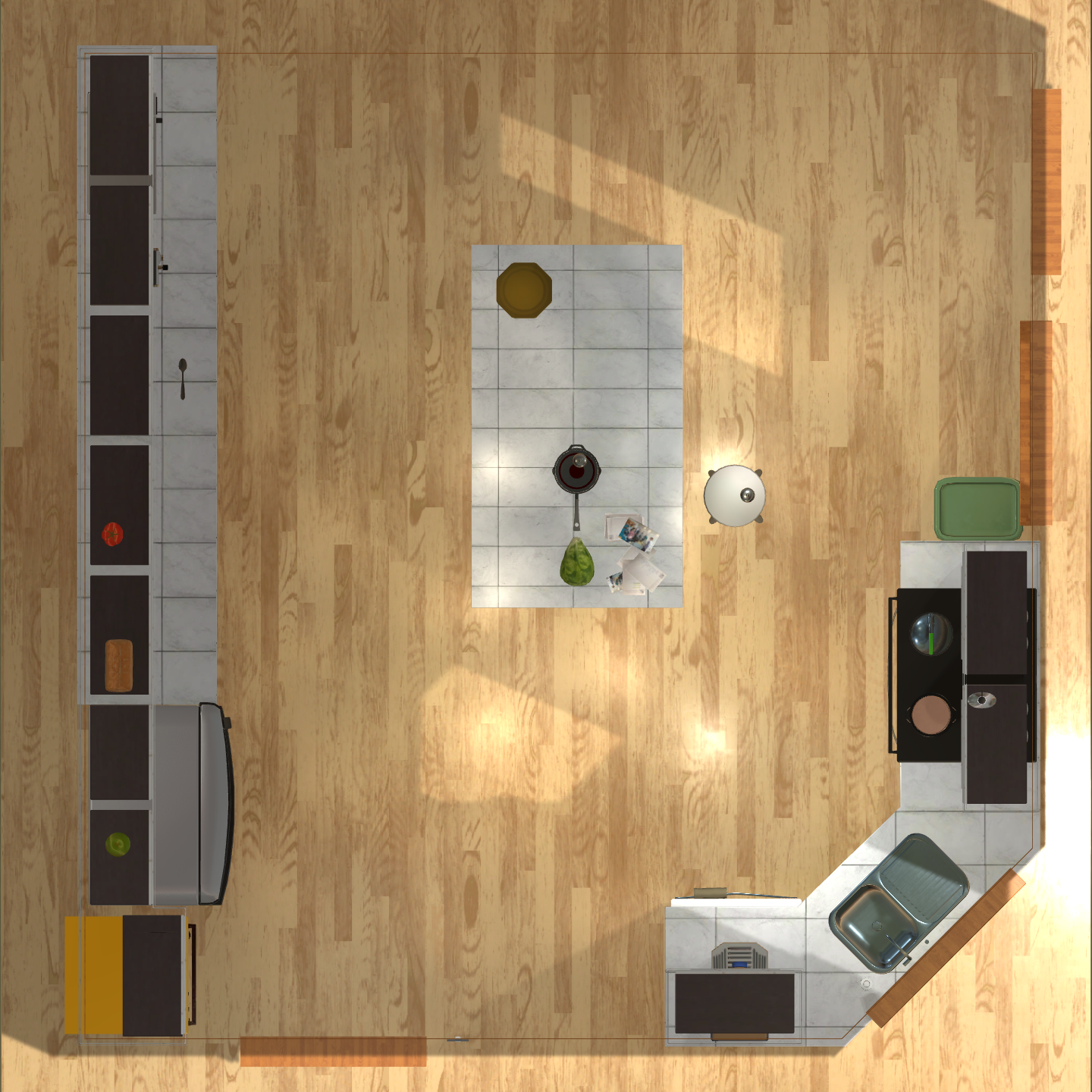}} \\
% \VarSty{ {\bf Object list:} } &\\
% \multicolumn{3}{p{0.97\columnwidth}}{CellPhone, Pot, GarbageBag, Shelf, SideTable, Plate, SprayBottle, Faucet, CoffeeMachine, Spatula, Toaster, StoveKnob, Pen, Sink, Ladle, Tomato, PaperTowelRoll, Book, Bread, CounterTop, Knife, Statue, SoapBottle, Vase, Cup, GarbageCan, Fridge, LightSwitch, SaltShaker, Bowl, HousePlant, Cabinet, Mug, Curtains, Apple, CreditCard} \\
\VarSty{ {\bf Instruction:} } & \multicolumn{2}{p{0.5\columnwidth}}{\VarSty{ {\bf Instruction:} }}\\
Can you clean the sink and the toilet, please? & \multicolumn{2}{p{0.5\columnwidth}}{Can you make me a sandwich?}\\
\VarSty{ {\bf Task planning:} } & \multicolumn{2}{p{0.5\columnwidth}}{\VarSty{ {\bf Task planning:} }}\\
Step 1: Grasp a sponge & \multicolumn{2}{p{0.5\columnwidth}}{Step 1. Grasp a plate}\\ 
Step 2: Move to the sink & \multicolumn{2}{p{0.5\columnwidth}}{Step 2. Grasp the knife}\\
Step 3: Wet the sponge & \multicolumn{2}{p{0.5\columnwidth}}{Step 3. Grasp a piece of bread}\\
Step 4: Scrub the sink & \multicolumn{2}{p{0.5\columnwidth}}{Step 4. Move the knife to the bread and slice it}\\
Step 5: Rinse the sponge & \multicolumn{2}{p{0.5\columnwidth}}{Step 5. Grasp another piece of bread}\\
Step 6: Grasp a towel & \multicolumn{2}{p{0.5\columnwidth}}{Step 6. Move the knife to the bread and slice it}\\
Step 7: Dry the sink & \multicolumn{2}{p{0.5\columnwidth}}{Step 7. Grasp a lettuce}\\
Step 8: Move to the toilet & \multicolumn{2}{p{0.5\columnwidth}}{Step 8. Tear the lettuce and place it on the plate}\\
Step 9. Grasp a scrub brush & \multicolumn{2}{p{0.5\columnwidth}}{Step 9. Grasp a tomato}\\
Step 10. Scrub the toilet bowl & \multicolumn{2}{p{0.5\columnwidth}}{Step 10. Slice the tomato and place it on the plate}\\
Step 11. Place the scrub brush back in its place & \multicolumn{2}{p{0.5\columnwidth}}{Step 11. Move the two slices of bread to the plate} \\
% Step 8. Grasp a bowl & \multicolumn{2}{p{0.5\columnwidth}}{\ \ \ \ \ \ \ \ \ \ \ \ \ \ with the sandwich}\\

    \end{tabular}
\end{tcolorbox}
\vspace{-2mm}
\caption{Examples of the generated multimodal triplet data including visual scenes, instructions and the corresponding plans.}
    \label{tab:example}
\end{minipage}
\vspace{-0.3cm}
\end{table*}

\subsection{Grounding Task Plans to Surrounding Scenes}
% In this section, we demonstrate how to train TAPA for embodied task planning. Then we show the whole pipeline of TAPA when apply it in an embodied environment.
% \subsubsection{Training}
% \subsubsection{Inference}
In order to ground the embodied task plan to the physical world with feasibility constraints, it is necessary to accurately obtain the object list in the scene without instance missing or false positives. We generalize the open-vocabulary object detector for object list acquisition since novel objects unseen in detector training may appear in the deployment scenarios. As shown in Figure \ref{fig:pipeline}, the agent collects RGB images in different locations to perceive the visual scenes to discover existed objects. We design several image collection strategies to explore the surrounding 3D scenes. The location selection criteria contains traversal positions, random positions, the overall center point and block-wise center points, and the agent rotates the camera to obtain multi-view images for each location selection criteria. Therefore, we formally write the image collection strategies $\mathcal{S}$ in the following:
\begin{equation}
    \mathcal{S}=\{(x,y,\theta)|(x,y)\in L(\bm{\lambda},\mathcal{A}), \theta = k\theta_0\}
\end{equation}where $(x,y,\theta)$ represents the location and camera orientation. $L(\bm{\lambda},\mathcal{A})$ means the location selection criteria with the hyperparameter $\bm{\lambda}$ and all sampled locations are required within the achievable area $\mathcal{A}$. The unit angle for camera rotation is set to $\theta_0$, and $k$ is an integer so that the agent collects visual clues in different directions of the scene. The hyperparameter that all location selection criteria share is the grid side length, where we divide the achievable area into grids. Traversal positions choose all grid points for RGB image collection. Random positions only randomly selected part of the grid points for visual information perception, and the hyperparameters also contain the ratio of sampled grid points. The overall center point stands for the center of the whole scene without any hyperparameters. The block-wise center points aim to choose the center of each division in the scene to efficiently acquire fine-grained visual information. Inspired by \cite{jiang2020site, liu2012site}, clustering methods can effectively divide the entire scene into several sub-regions to improve the performance of perception, so that the prior information of the room layout is embedded into the image collection strategy with the K-means clustering method. Meanwhile, we employ within cluster sum of squared errors (WCSS) principle to select the optimal number of clusters for each scene. Compared to the images collection strategy of traversal points, the block-wise center point only traverses centroids of the subregions to acquire sufficient visual information.

%is first initialized randomly in an unknown environment $X_s$ with instruction $X_q$ from the user. The $X_s$ will be sampled uniformly according to the value of grid size (e.g., 0.25, 0.75, etc.), and each point $P = \left\{(x_1, y_1), (x_2, y_2), ..., (x_m, y_m)\right\}$ obtained represents a location that the robot can reach, as shown by the colored dots in Figure \ref{fig:data_generation}. The grid size of 0.25 and 0.75 both represent the robot traversing the whole scene with mesh size. At each point, the robot will perform a surround-view perception to obtain $X_v$.

The embodied task planner requires the information of all existed objects in the scene to generate executable action steps, where we generalize the open-vocabulary object detector to the collected multi-view RGB images for the object list acquisition. The predicted object list $\hat{X}_l$ for the scene is acquired by removing the duplicated object names in the detection results of multi-view images:
\begin{equation}
\hat{X}_l={\rm Rd}\big(\bigcup_{i} D(\bm{I}_i)\big)
\end{equation}where $\rm{Rd}$ is the operation of removing duplicate object names and $D(\bm{I}_i)$ represent the detected object names for the $i_{th}$ RGB image collected in the scene. With our inference prompt $P_{in}$ shown in Table \ref{tab:prompt} of the supplementary material, the human instruction $X_q$ and the predicted object list $X_l$ are considered in our TaPA to generate the executable action plans $X_a$:
\begin{equation}
    X_a={\rm TaPA}(P_{in}, \hat{X}_l, X_q)
\end{equation}By combining the perception results of existed objects $\hat{X}_l$ with the instructions $X_q$, TaPA will give the executable action sequence $X_a$ to complete the requirements of $X_q$ according to the realistic scene constraint. According to our empirical study, we chose the block-wise center point for multi-view RGB image collection. The grid size in our location selection criteria is set to 0.75 and the unit angle for camera rotation is $2\pi/3$.

%% file: tex/experiment.tex
In this section, we conduct extensive experiments with our generated multimodal dataset where the visual scenes come from the simulator AI2-THOR. We first introduce the evaluation metric of the generated action plans. Then we compare our TaPA with the state-of-the-art LLMs and LMMs to show our superiority in embodied task planning. To further explore the effectiveness of different scene information embedding approaches, we evaluate various image collection strategies in our ablation study. We employ the LLaMA-7B pre-trained language model as the backbone of our task planner, which is finetuned with our generated multimodal dataset. The maximum token number of our task planner is set to 512, and we leverage the Detic open-vocabulary object detection framework to collect the information of existed objects. All experiments were accelerated by 8 GTX 3090 GPUs.

\subsection{Evaluation Metrics}

\begin{table}[t]
\centering
\caption{Comparison of different LLMs and LMMs on the task of embodied task planning. For the prompt of baseline methods, LLaMA and LLaVA both employ the same prompt in the their original finetuning phase, while GPT-3.5 adopts the same prompt of TaPA for multimodal data generation.}
\label{vs_baseline}
\vspace{0.2cm}
\begin{tabular}{c|c|c|c|c|c}
\hline
\makebox[0.13\textwidth]{Method} & \makebox[0.13\textwidth]{Kit.} & \makebox[0.13\textwidth]{Living.} & \makebox[0.13\textwidth]{Bed.} & \makebox[0.13\textwidth]{Bath.} & \makebox[0.13\textwidth]{Avg.} \\ \hline
LLaVA & 14.29 & 42.11 & 33.33 & 0.00 & 22.43 \\
GPT-3.5 & 28.57 & 73.68 & 66.67 & 50.00 & 54.73 \\
LLaMA & 0.00 & 10.52 & 13.33 & 0.00 & 5.96 \\ \hline
TaPA & 28.57 & 84.21 & 73.33 & 58.33 & 61.11 \\ \hline
\end{tabular}
\vspace{-0.2cm}
\end{table}

%In order to demonstrate the performance of the proposed method clearly, we aim to illustrate the performance of the proposed method in following instructions on robot manipulation utilizing quantitative measures of human scoring.
% In addition, the proposed data generation pipeline is employed to gather instruction-following samples for the 20 scenes in the validation set. 
% These samples serve as a reference baseline for scoring purposes.
For the deployment of our TaPA, we feed the instructions and the predicted object list in the scene for the task planner to generate the action steps. We hired 30 researchers in large multimodal models as volunteers to vote for the success of the generated action plans, and each generated action plan is evaluated by three volunteers. The volunteers are shown with the groundtruth object list of each scene, the instruction and the generated action plans, where the volunteers should judge whether implementing the action steps can successfully completes the instructions. There are two types failure cases including counterfactuals and hallucination. Counterfactuals indicate that the plans violate the physical rules in the real world (e.g. grasping the doorknob before moving to the door), and hallucination means the action plans require the agent to interact with objects that do not exist in the scene. An exceptional case is that the interacting objects can be part of the object existed in the scene  (e.g. trash can lid and trash can) or a synonym of an object (e.g. mug and cup). The generated action plans are considered to be successful if at least two volunteers out of three regarding the steps can be implemented to satisfy the human instruction. The volunteers were also requested to annotate the type of failure for the unsuccessful cases. We report the ratio of successful cases for different scene types and plan generation models.

%We feed the generated instructions with the object categories detected by Detic into the proposed method to generate the parsed actions. After obtaining the responses of each model, we present them to the volunteers with \textbf{Input\_GT}, a list of all the objects in the room, and the people judge whether the instruction can be successfully executed based on the listed execution actions and the objects in the room, to count the instruction execution success rate. Specifically, each instruction sample is scored by three volunteers and is judged to be successful if no less than two people consider the instruction to be successfully executed. Volunteers were requested to jointly judge the samples from both the hallucinatory and counterfactual perspectives. The counterfactual perspective mainly required that the responses could not contain actions that violated general life common sense (e.g., grasping the doorknob before moving to the door) and could complete the instructions as requested.  As for the hallucinatory perspective, each step is carefully examined to determine whether it interacts with an object that is not present in the room (the interacting object is not in \textbf{Input\_GT}). Note that the interacting object can be a synonym of an object in \textbf{Input\_GT} or a part of it (e.g. mug and cup, trash can and trash can lid).
% Meanwhile, volunteers are also required to give relative scores in combination with the baseline results to judge the performance of our method compared to GPT-3.5.

\begin{table}[t]
\centering
\caption{The average execution success rate of generated action steps for different RGB image collection strategies in scene perception. $G$ represents the side length of grids in location selection. $\Delta \theta$ represents the unit angle of camera rotation. $N$ represents the ratio of randomly selected points compared to all grid points in achievable area.}
\label{Success_rate} 
% \resizebox{\linewidth}{!}{
\vspace{0.2cm}
\begin{tabular}{cl|c|c|c|c|c|c}
\hline
\multicolumn{2}{c|}{Strategy and Parameters} & \makebox[0.07\textwidth]{\#Images} & \makebox[0.07\textwidth]{Kit.} & \makebox[0.07\textwidth]{Living.} & \makebox[0.07\textwidth]{Bed.} & \makebox[0.07\textwidth]{Bath.} & \makebox[0.07\textwidth]{Avg.} \\ \hline
\multicolumn{1}{c|}{\multirow{4}{*}{Traversal}} & G=0.25, D=60 & 782.4 & 14.29 & 73.68 & 46.67 & 33.33 & 41.99 \\
\multicolumn{1}{c|}{} & G=0.25, D=120 & 391.2 & 14.29 & 73.68 & 53.33 & 50.00 & 47.83 \\
\multicolumn{1}{c|}{} & G=0.75, D=60 & 80.7 & 28.57 & 73.68 & 46.67 & 33.33 & 45.56 \\
\multicolumn{1}{c|}{} & G=0.75, D=120 & 40.4 & 14.29 & 63.16 & 60.00 & 41.67 & 44.78 \\ \hline
\multicolumn{1}{c|}{\multirow{4}{*}{\begin{tabular}[c]{@{}c@{}}Random\\ (G=0.75)\end{tabular}}} & N=1\%, D=60 & 6.0 & 28.57 & 78.95 & 26.67 & 50.00 & 46.05 \\
\multicolumn{1}{c|}{} & N=1\%, D=120 & 3.0 & 21.43 & 73.68 & 46.67 & 50.00 & 47.95 \\
\multicolumn{1}{c|}{} & N=75\%, D=60 & 63.0 & 35.71 & 73.68 & 53.33 & 25.00 & 46.93 \\
\multicolumn{1}{c|}{} & N=75\%, D=120 & 31.5 & 28.57 & 73.68 & 53.33 & 33.33 & 47.23 \\ \hline
\multicolumn{1}{c|}{\multirow{2}{*}{\begin{tabular}[c]{@{}c@{}}Layout Priori\\ (G=0.75,D=60)\end{tabular}}} & Overall Center & 6.0 & 28.57 & 68.42 & 33.33 & 58.33 & 47.16 \\
\multicolumn{1}{c|}{} & Partial Center & 23.1 & 28.57 & 84.21 & 73.33 & 58.33 & 61.11 \\ \hline
\end{tabular}
% }
\vspace{-0.3cm}
\end{table}

\subsection{Experimental Results}
% baseline

In this section, we compare our TaPA method with the state-of-the-art LLMs including LLaMA and GPT-3.5 and LMMs containing LLaMA on 60 validation samples, and the success rate of the generated action steps from different methods are shown in Table \ref{vs_baseline}.
TaPA achieves optimal performance among all large models on all four scenes including kitchen, living room, bedroom and bathroom, and the average success rate of TaPA is 6.38\% (61.11\% vs. 54.73\%) higher than GPT-3.5 on the task of embodied task planning after instruction finetuning.

Since agents in kitchen scenes usually deal with complex cooking instructions in more steps, the performance of current large models is lower than in other room types. Meanwhile, the poor performance of LLaVA reflects the fact that the overall scene information cannot be represented by a single image in the visual question answering task, and the insufficient scene information leads to a low success rate of task planning. The success rate of LLaMA is far below other methods, which even cannot succeed in completing tasks in the kitchen and bathroom scenes without instruction finetuning. Figure \ref{fig:data_comparison} illustrates the percentage of failure cases in embodied task planning for different large models. Counterfactuals represent that the generated actions violet the physical rule in the realistic world, and hallucinations mean the actions aim to interact with objects that are not in the scene. TaPA is embedded with more expert knowledge in embodied task planning after instruction finetuning, which has the lowest percentage of counterfactual occurrences. Moreover, TaPA can better understand the list of input objects, with a 26.7\% (40.0\% vs. 13.3\%) and 5.0\% (18.3\% vs. 13.3\%) decrease in the percentage of hallucination cases compared to LLaVA and GPT-3.5 respectively.

% \begin{figure}[t]
% \centering
% \includegraphics[width=1.0\textwidth]{figs/figure_5_v2.png}
% \caption{Input scene with TaPA, Llava, GPT-3.5 and Llama output examples. Compared to other approaches, TaPA output results are not counterfactual and hallucinatory, and the generated plans can be executed.}
% \vspace{-0.3cm}
% \end{figure}
% height=5.2cm
% width=0.8\textwidth
% 
\begin{table}
  % \begin{minipage}{0.99\textwidth}
\centering
\resizebox{\linewidth}{!}{
\begin{tabular}[t]{llll}
\toprule
 \multicolumn{4}{l}{\bf Qualitative results, Living room:}  \\
\midrule
\multicolumn{4}{c}{ \includegraphics[height=5.2cm]{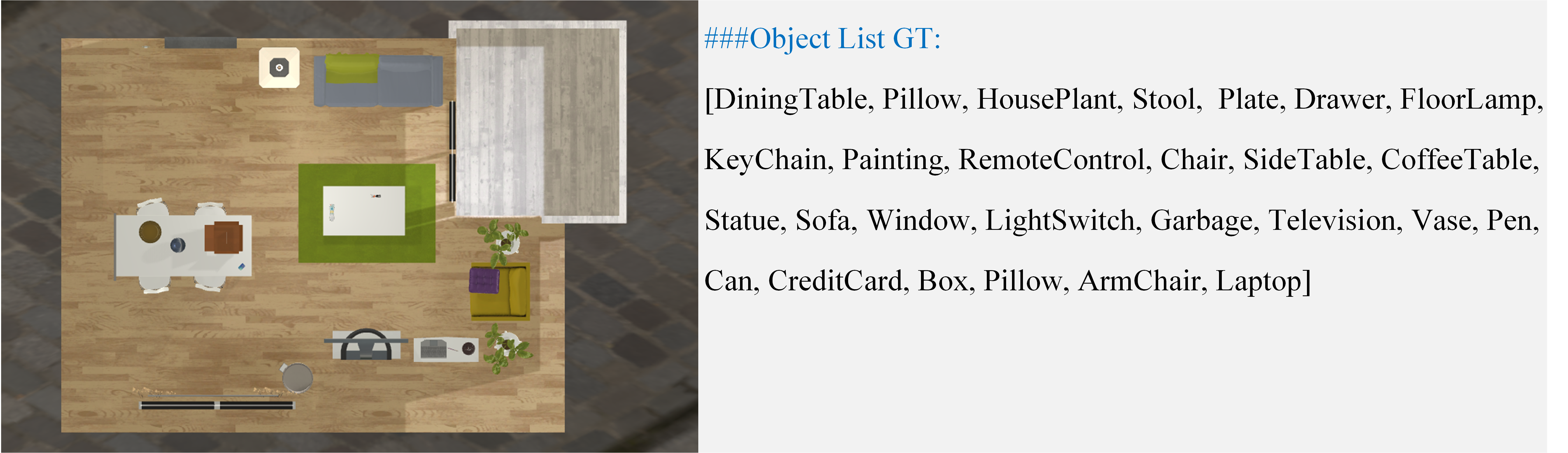}} \\
\midrule
User & \multicolumn{1}{l|}{I want to watch TV} & User & I want to watch TV \\
TaPA & \multicolumn{1}{l|}{\begin{tabular}[t]{@{}l@{}}Step 1. Move closer to the   television set\\      Step 2. Grasp the TV remote control\\      Step 3. Turn on the TV\\      Step 4. Adjust the volume to a suitable level\\      Step 5. Select a TV show or movie\\      Step 6. Watch TV\\      
%Step 7. Done
\end{tabular}} & GPT-3.5 & \begin{tabular}[t]{@{}l@{}}Step 1. Move closer to the TV   \\      Step 2. Grasp the remote control \\      Step 3. Rotate the remote control to point at the TV\\      Step 4. Press the power button to turn on the remote control \\      
%Step 5. Done
\end{tabular} \\ \midrule

User & \multicolumn{3}{l}{I want to watch TV} \\
LLaVA & \multicolumn{3}{l}{\begin{tabular}[t]{@{}l@{}}
1.Locate the TV remote control, which is typically placed on a table or mounted on the wall near the TV.\\ 
2.Point the remote control at the TV and press the power button. The power button is often located on \\
the top or bottom of the remote control.\\
3.If the TV doesn't power on immediately, press the channel or input button to select the correct input source. \\
The input button is usually located on the right side of the remote control or may be a menu button that allows \\
you to access the settings.\\
4.The TV should now turn on and display the channel or content you have selected.
\end{tabular}} 
\\ \midrule

User & \multicolumn{3}{l}{I want to watch TV} \\
LLaMA & \multicolumn{3}{l}{\begin{tabular}[t]{@{}l@{}}
Sit in a comfortable chair, place a pillow on the floor, and turn on the TV. \\
Open up the drawer to get a remote control and place the TV on a table. Adjust the volume on the speakers. \\
Locate the plug for the television and plug it in. Place the bowl of objects on the table next to the TV. \\
(Objects: chair, sofa, pen, figurine, table, dog, lampshade, doorknob, bed, toy, drawer, person, statue\_(sculpture), \\
flowerpot, stool, monitor, (computer equipment), speaker, computer\_monitor, ……)
\end{tabular}}
\\ \toprule

% Step 2. Grasp the TV remote control  & Step 2. Grasp the remote control\\
% Step 3. Turn on the TV  & Step 3. Rotate the remote control to point at the TV\\
% Step 4. Adjust the volume to a suitable level & & Step 4. Press the power button to turn on the remote control \\
% Step 5. Select a TV show or movie \\
% Step 6. Watch TV \\
% Step 7. Done
\end{tabular}
}
\vspace{0.3cm}
\caption{Visualization of generated action plans by different large models. The inputs of LLaMA and GPT-3.5 are the predicted object list by our visual perception module, while LLaVA generates the action steps only based on one single image.}
\label{tab:vis_results}
  % \end{minipage}
\vspace{-0.5cm}
\end{table}

%Specifically, we implement different navigation strategies by modifying the minimum (grid size) and the minimum self-centered rotation angle (degree-step) of the robot in the scene.
% 
%Nevertheless, the computational burden associated with fully exploring the scene is difficult to justify. 

We also investigate the effectiveness of different image collection strategies that perceive the scene information by acquiring the list of existed objects. Specifically, we employ location selection criteria including random positions, traversal positions, the overall center point and block-wise center points with various hyperparameters containing the grid size and the sampling ratio in random positions, and we also change the unit angle for camera rotation. The success rate of different image collection strategies is demonstrated in Table \ref{Success_rate}. We also show the number of collected images for various criteria to reveal the collection and computational cost. For the traversal positions, reducing the grid size significantly increases the image collection and the computational cost due to the numerous RGB images, while the average success rate remains similar (47.83 vs. 44.78) because the large grid size can collect images with sufficient information of the small-scale scenes from AI2-THOR. Similar reasons result in the phenomenon for random positions that increasing the sampling ratio and reducing the unit angle for camera rotation by collecting images in more locations cannot boost the success rate (47.95 vs. 47.23, 46.93 vs. 47.23). Since the traversal positions with small grid sizes (G=0.25) collects extremely large number of images, decreasing the unit angle for camera rotation significantly decreases the success rate because the redundant object list degrades the planning capacity of LLMs.

\begin{wrapfigure}{r}{6.5cm}%
\centering
\vspace{-0.3cm}
\includegraphics[width=0.47\textwidth]{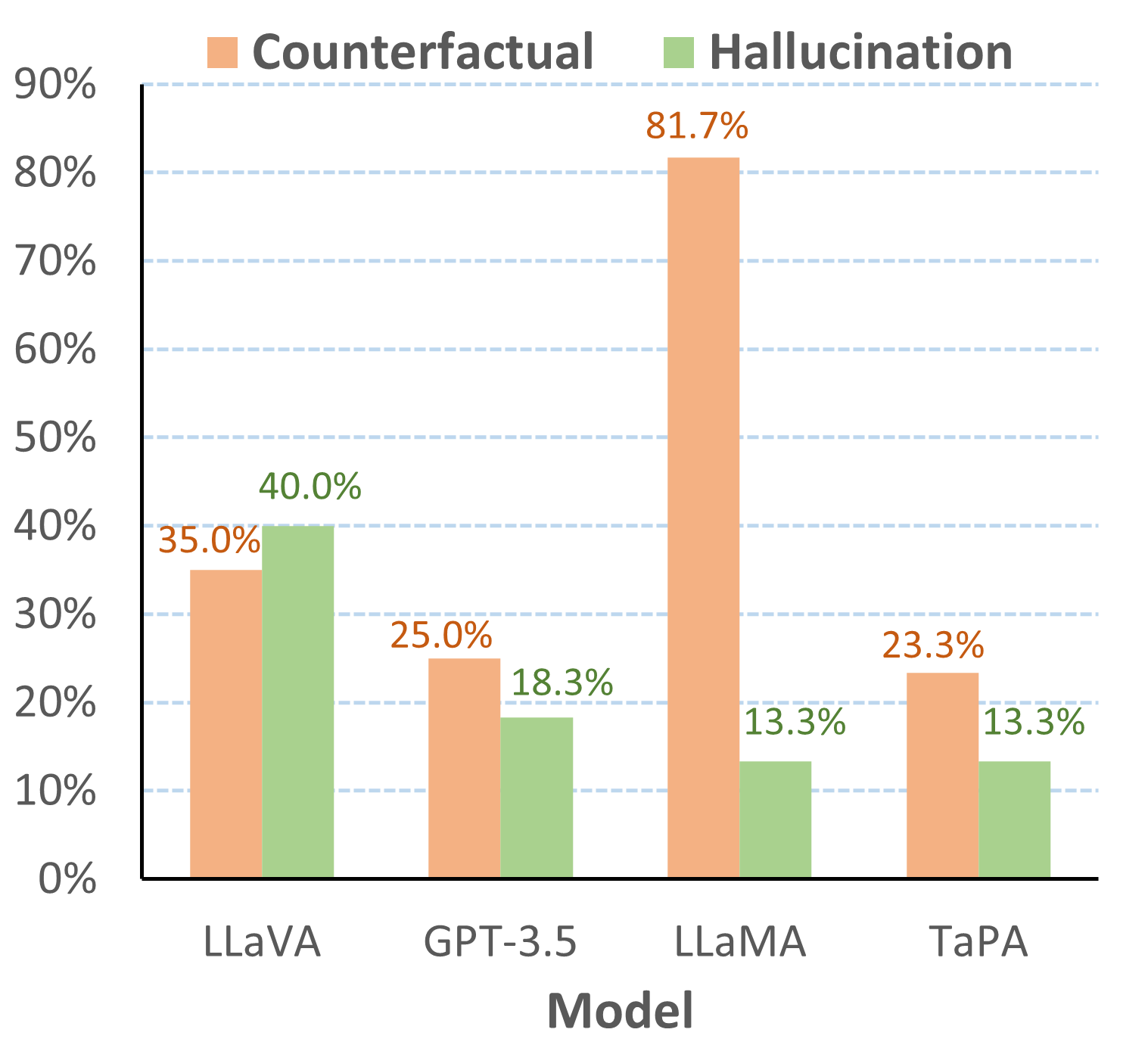}
\caption{The percentage of different failure cases in embodied task planning for various large models.}
%Counterfactuals represent generated actions that do not complete instructions or go against the common sense of life, and hallucinations represent sequences of actions that have objects that are not in the scene
\vspace{-0.1cm}
\end{wrapfigure}

Comparing all location selection criteria, block-wise center points achieve the highest success rate because of the effective representation of the existed objects in the scene. Block-wise center points observe the scene with the high coverage rate, while only a few RGB images are collected for scene representation. Therefore, sufficient scene information is captured by the acquired object list without redundancy. The performance of random positions and the overall center point is similar because the scale of scenes in AI2-THOR is small and one image collection location can also receive sufficient information. The traversal positions obtain the lowest success rate since collecting excess images lead to the higher probability of false positives in open-vocabulary object detection, which degrades the success rate because of the redundant object list.

Among all room types, the success rate in the kitchen scenes is the lowest since the instruction for kitchen tasks (e.g. sandwich making) usually requires long plans with much more action steps. With the increase of the interacted objects in the task plan, the probability of hallucination is higher so that the plans are more likely to fail. On the contrary, the success rate of tasks in the living rooms is high due to the simple instructions (e.g. turning off lights). By observing the success rate of kitchen tasks across different location selection criteria, false positives in object detection that usually appear in traversal location selection criteria degrade the performance most significantly. Since the object list is redundant, the complex tasks in kitchen scenarios are more prone to the noise in the object list.

We also show an example of generated action steps from different large models for the given scene in Table \ref{tab:vis_results}. The scene is demonstrated in the top-down view, and we also provide the groundtruth object list for reference. The content from LLaMA is irrelevant to the human instructions, while LLaVA provides plans that are not executable due to the non-existed objects. Although GPT-3.5 can also yield plausible embodied task plans, the action steps from our TaPA are more complete and more consistent with human values.

%To further elucidate the efficacy of the proposed approach, we provide visualization results demonstrating the performance of different baseline models for parsing user instructions.

%Compared to the results of LLaVA and LLaMA parsing, our TaPA generates plans with higher executability. Compared to GPT-3.5, the results of TaPA parsing are more consistent with human values and provide more complete actions.
%These visualization results encompass various elements, including the selection of cruise positions, segmentation outcomes highlighting the instances within the captured images, the input commands given, and the corresponding execution actions parsed by our method. 
%These visualizations serve as a valuable reference for readers to better comprehend the performance of our approach.

%% file: tex/conclusion.tex
In this paper, we have presented a task planning agent called TaPA for embodied task planning, where the executable action steps are generated for subsequent robot navigation and manipulation to complete human instructions. We first construct a multimodal dataset where each sample is a triplet including the visual scenes, the instructions and the corresponding plans. The dataset is generated with GPT-3.5 by considering the list of all objects in the scene and the designed text prompt, which is leveraged to tune the instruction model to generate executable actions. For inference, we collect multi-view RGB images in different achievable locations, and leverage an open-vocabulary object detection framework to discover the object list of the scene for the finetuned instruction model. The statistics of our collected multimodal dataset indicate that our tasks are much more complex than conventional benchmarks on instruction-following tasks with longer implementation steps, and the extensive evaluation results show that our TaPA outperforms the state-of-the-art LLMs and LMMs on the plausibility of generated action plans. 

%% file: tex/supp.tex
\begin{table*}[t!]\centering
\renewcommand\arraystretch{1}
\begin{minipage}{1.0\columnwidth}\vspace{0mm}    \centering
\begin{tcolorbox} 
    \centering
   
    %  \hspace{-10mm}
      \footnotesize
    \begin{tabular}{p{0.97\columnwidth} c}
    
\VarSty{ {\bf Rule description:} } &\\
You are an indoor service robot named Garybot and you are inside a room. What you see is provided with a list of objects that contains all the objects in the room you are in. The location of the objects in the list you are guided in advance, without reasoning about the spatial relations of the objects. Execute all the instructions as you are located in the room.&\\
\\
Design a conversation between you and the person you are serving in the room. The answer should be the tone of the service robot located in the room and performing the action specifically. 
The generated instructions can be described in different tones. Ask for various instructions and give the corresponding series of actions with a maximum of 15 steps. & \\
\\
Only include instructions for their corresponding actions only utilizing atomic motions (Grasp, Release, Lift, Place, Rotate, Push, Pull, Align, Press, Pour, Move): &\\
(1) Generate operation instructions using only the objects in the list with the actions that must be performed to complete the operating instructions;&\\
(2) Do not generate any instructions or actions that cannot be executed with confidence;&\\
(3) Do not generate any instructions or actions with (Target: \VarSty{[Object]}), \VarSty{[Object]} is outside the list of objects.\\
\\
Again, the object being manipulated cannot be located outside the list. 
Please double-check that Target: \VarSty{[Object]} is in the list at each step and that \VarSty{[Object]} is in the list.
When evaluating the existence of \VarSty{[Object]}, consider its original part or component, its function, and whether it can be replaced by an object in the list, and if it is satisfied, you can iterate over each element in the list to find an alternative and replace \VarSty{[Object]}. \\
\\
\VarSty{ {\bf Few-shot samples:} } & \\
List of objects: [wine, cup, glass, remote control, TV, table, desk, chair] \\
Generate the instruction: Give me a drink \\
Necessary actions: \\
Step 1. Grasp a bottle of wine (Target: wine)\\
Step 2. Grasp a glass (Target: bowl)\\
Step 3. Place the cup on the table (Target: glass, table)\\
Step 4. Pour the wine into the glass (Target: wine, glass)\\
Step 5. Grasp the glass with wine (Target: glass)\\
Step 6. Move to the person and hand over it\\
Step 7. Done\\
\\
Generate the instruction: Please turn on the TV\\
Necessary actions:\\
Step 1. Grasp the remote control (Target: remote control)\\
Step 2. Move closer to the TV (Target: TV)\\
Step 3. Rotate the remote control to point at the TV (Target: remote control, TV)\\
Step 4. Press the power button to turn on the remote control (Target: remote control)\\
Step 5. Done\\

    \hrulefill & \\
\\
   \VarSty{ {\bf Prompt for training and inference:} } & \\
Below is an instruction that describes a task, paired with an input that provides further context.
Write a response that appropriately completes the request.\\
Instruction: \textcolor{green}{$X_q$}. Input: \textcolor{green}{$X_l$}. Response: \textcolor{orange}{$X_a$}.\\

    \end{tabular}
\end{tcolorbox}
\vspace{-2mm}
\caption{Our prompt for multimodal dataset generation (upper) and training/inference of TaPA (bottom). \textcolor{orange}{$X_a$} is empty unless the prompt serves as a ground-truth.}
    \label{tab:prompt}
\end{minipage}
\end{table*}

% 增加prompt的解释
The prompts utilized to generate the instruction-following dataset from GPT-3.5 are illustrated in Table \ref{tab:prompt}. Specifically, we set a specific work scene for GPT-3.5 and indicate the need to generate instructions and corresponding actions by the agent itself.
We also set the rules to constrain the instructions generated by GPT-3.5 with improved executability and confidence. Meanwhile, we require GPT-3.5 to add an additional (Target: \VarSty{[Object]}) query to each generated action to further check for hallucinations. If the interacting object in the generated plan is not in the input $X_l$, it is necessary to check if there is an alternative item to replace it. 
An exceptional case is that the interacting objects can be part of the existing objects or a synonym of an object. We also provide some examples to standardize the form of the input and output for the generated data.